\documentclass{article}

\usepackage[preprint, nonatbib]{neurips_2021}

\usepackage[utf8]{inputenc} 
\usepackage[T1]{fontenc}    
\usepackage{hyperref}       
\usepackage{url}            
\usepackage{booktabs}       
\usepackage{amsfonts}       
\usepackage{nicefrac}       
\usepackage{microtype}      
\usepackage{xcolor}         
\usepackage{graphicx}
\usepackage{amsmath}
\usepackage[numbers]{natbib}

\title{Quantitative probing: Validating causal models with quantitative domain knowledge}

\author{%
  Daniel Grünbaum \\
  ams OSRAM Group\\
  University of Regensburg \\
  Regensburg, Germany \\
  \texttt{daniel.gruenbaum@ams-osram.com} \\
   \And
   Maike L. Stern \\
   ams OSRAM Group\\
   Regensburg, Germany \\
   \And
   Elmar W. Lang \\
   Department of Physics \\
   University of Regensburg \\
   Regensburg, Germany
}
\begin{document}

\maketitle

\begin{abstract}
  We present quantitative probing as a model-agnostic framework for
  validating causal models in
  the presence of quantitative domain knowledge.
  The method is constructed as an
  analogue of the train/test split in correlation-based machine learning and
  as an enhancement of current causal validation strategies that are consistent with
  the logic of scientific discovery.
  The effectiveness of the method is
  illustrated using Pearl's sprinkler example, before a thorough
  simulation-based investigation is conducted.
  Limits of the technique are
  identified by studying exemplary failing scenarios, which are furthermore used
  to propose a list of topics for future research and improvements of the
  presented version of quantitative probing.
  The code for integrating quantitative probing into causal analysis,
  as well as
  the code for the presented simulation-based studies of the effectiveness of
  quantitative probing is provided in two separate open-source Python packages.
\end{abstract}

\section{Introduction}
The topic of causal inference has been a focus of extensive research in statistics,
economics and artificial intelligence.
\cite{causality_pearl, angrist_imbens_rubin_instrumental,
rubin_potential_outcomes, causation_prediction_search, peters_elements_of_causal_inference,
holland_statistics_and_causal_inference}
In contrast to more traditional
statistics, methods of causal inference allow predicting the behavior of a system not only in
situations where we passively observe certain evidence,
but also in situations where we actively intervene in the data generating process.
Prior to the rise of causal inference,
ruling out confounding in such predictions required performing costly and possibly
harmful randomized controlled trials \cite{kendall_rct}.
Causal inference, on the contrary, provides the methodology to infer the effect of hypothetical
actions from passively collected observational data and additional assumptions, which can be
encoded in graphs
\cite{causality_pearl}
or conditional independence statements \cite{rubin_potential_outcomes}.
The benefits of evaluating a wide range of possible actions without actually having
to perform them are obvious in fields including medical treatment, industrial
manufacturing or governmental policy design.
However, in contrast to correlation-based prediction techniques,
such as linear regression models,
support vector machines
or neural networks,
which can all be validated using the well-known train/test split \cite{hastie_esl},
the challenge of validating the causal models responsible for the predictions of
interest is still unsolved.
Without having validated the causal model, the promising perspective of predicting the behavior
of a system under interventions from purely observational data is compromised by our uncertainty
about the validity of the underlying model.
This gap in the methodology needs to be filled, before decision makers in the above-mentioned
application domains can leverage the powerful methods of causal inference, in order to complement,
enhance or even replace the
costly current method of randomized controlled trials.
In this paper, we present quantitative probing, a novel approach for validating
causal models, which applies the logic of scientific discovery \cite{popper}
to the task of causal inference.
Both, the code for integrating quantitative probing into causal analysis and the
code for simulation-based studies of the effectiveness of quantitative probing
are provided in two separate open-source Python packages
\cite{cause2e_github,qprobing_github}.
The article is structured as follows:
Chapter \ref{iid} reviews the train/test split as the classical method for validating
correlation-based statistical models, and explains why it is not possible to
directly transfer the method to the field of causal models.
Chapter \ref{sota} presents different approaches to the validation of causal models and
discusses the limitations of the currently available validation techniques.
Chapter~\ref{causal_models} briefly clarifies the notion of a causal model and introduces
causal end-to-end analysis as a model that generalizes other types of
causal models.
Chapter \ref{qprobing} synthesizes the idea of quantitative probing from the observations
in the previous chapters, and relates the approach to the established method of
scientific discovery.
The concept is illustrated using Pearl's well-known Sprinkler example \cite{causality_pearl}
and
assumptions are explicitly stated, in order to define the current scope of the
quantitative probing approach.
Chapter \ref{results} provides simulation-based evidence for the effectiveness of
quantitative probing as a validation technique. Special cases where the method
fails to detect misspecified models are investigated in detail, in order to
identify limitations and future enhancements.
Chapter \ref{conclusion} concludes the article by summarizing the main points and proposes
concrete questions for future research.

\section{The role of the i.i.d. assumption in model validation}\label{iid}
In order to gauge the difficulty of the challenge that validation poses for
causal inference procedures,
it is worth taking a step back and recapitulating the reasoning behind the
predominant strategy for validating traditional correlation-based statistical
learning methods.
These methods, which encompass both classification and regression algorithms,
such as linear regression,
support vector machines
or numerous variations of deep learning with neural networks,
have one crucial assumption in common \cite{hastie_esl}:
Every sample that we have observed in fitting the model, as well as every sample
that we will need to feed into the final model for classification or regression,
is drawn from the same distribution, and they all are drawn independently of
each other.
This assumption is
commonly referred to as the \emph{i.i.d. assumption},
which stands for \emph{independent and identically distributed}.
However, it is
seldomly explicitly mentioned because of how deeply ingrained it is in all
correlation-based thinking about machine learning. The two parts of the term
i.i.d. have important consequences for how we train or validate machine learning
models:

The independence assumption is the implicit foundation of the current practices
for model training:
If we did not assume that all samples are drawn independently from each other,
the likelihood
\begin{equation}
  p(X,Y|\theta) = \prod_{i=1}^n p(x_i, y_i| \theta)
\end{equation}
of observing samples with features $X$ and labels $Y$ for a given parameter $\theta$
would not factorize over all the samples $(x_i, y_i)$.
The consequence would be that the commonly used error metrics, e.g. the
mean squared error or mean absolute error, would lose their theoretical
backing: All of them are based on summing up independently computed prediction
errors of the model for each sample, which is justified precisely by the
factorization property of the likelihood (or equivalently the summation property
of the log-likelihood).

The assumption of identical distribution enables the use of the
train/test split for model validation:
If all the samples that we will ever need to classify stem from the same
distribution as the observed labelled data, we can fit our model on the observed
data and be confident that the obtained model
will also be suitable for classifying the new incoming data. Even the
thereby caused risk of overfitting, i.e. learning overly specific
characteristics of the training data that fail to generalize and lead to a worse
performance on new data, can be mitigated using the same distributional
assumption: If we do not train the model on all the labelled data, but only on a
subset of it (the \emph{training set}), we can evaluate its performance on the
rest of it (the \emph{test set}), given that the correct labels for the test set
are available to us. Additionally, we have not used the test set for model training and
it is statistically identical to the new data
that we will have to classify in the actual task, because it is drawn from the same distribution.
Therefore, the expected value
of the prediction error for any unseen sample is equal to the mean error on the
test set. In summary, we can base our confidence in the predictions of the model
on its performance on the test data, which is enabled precisely by the
assumption of identical distribution.

If we now try to transfer these techniques to the training and validation of
\emph{causal} models, we will inevitably face severe problems.
The observational data that we use to train our causal models can very well
follow the i.i.d. assumption.
The notion of using test samples from the same distribution to evaluate how well
the model performs on hypothetical queries, however,
is diametrically opposed to the
task of causal inference: We want to predict what happens under certain
interventions, and an intervention is precisely the act of changing the data
generating process. A change in the data generating process, of course, generally entails
a change in the distribution from which the samples are drawn.
Since we are given only observational data, we need to use something other than
the pure data to gauge the usefulness of our model for the task of predicting
the behavior of the system under interventions.

\section{State of the art}\label{sota}
In recent years, others have already tackled the challenging question of how to validate causal
models.
One common approach is to accept that the model is likely to have
flaws. If these can be identified and bounded, we can still use this knowledge
to obtain error bounds around the estimates of the model in the spirit of a
sensitivity analysis
\cite{jesson_quantifying_ignorance, cinelli_making_sense_of_sensitivity, chernozhukov_omitted_variable_bias,veitch_sense_and_sensitivity}
.
As an example, if we are unsure about the direction of one
edge in the causal graph, we can perform the identification and estimation of
the target effect for both versions of the causal graph and use the difference
of the resulting estimates as an uncertainty measure. A similar approach can be
used for all other parts of the causal analysis, such as choosing an estimator,
hyperparameters, or treatment and outcome models. These methods are easy to
implement, but they depend on the specifics of creating the causal model.
Another drawback is that they do not tell us how close our model is to the
correct one. Quite on the contrary, they expect us to know in which ways and how
far we have deviated from the true model, in order to plug these deviations into
the different variants of sensitivity analysis. For complex models, it is
unlikely for researchers to take care of all possible deviations, which then
leads us to doubt not only our candidate model, but also the sensitivity
analysis that was supposed to be the answer to these doubts.

A second stream of research
\cite{neal_realcause}
takes a more straightforward approach to dealing with the
unknown ground truth for the behavior of the system under interventions: We take the
same algorithm that is used to create our causal model, but instead of applying
it to our real data, we use simulated data. If the simulation environment can
also generate data for the interventional scenarios of interest, we can then
compare the predictions of the model to the simulated ground truth. While the
advantages of this convenient approach are obvious, it rests on several critical
assumptions that are hard to verify.
First, we cannot choose any simulated data at will to
gauge the performance of our model on the real data. It is clear that the data
generating process in the simulation scenario must be similar to the data
generating process in the actual scenario of interest. However, this requires
knowledge about the scenario of interest, which we might not have. If we fully
understood the data generating process, then we would not have to perform a
causal analysis in the first place. A second problem is that we are not directly
evaluating the performance of a model that we want to use for the real task, but
we are evaluating the performance of a different model that is trained on the
simulated data. The two models of interest are linked by the algorithm that is
used to create them, but it is not clear to which extent the performance of one
model will transfer to the performance of the other via this link.

A third approach
\cite{dowhy_paper}
tries to employ refutation tests to probe candidate models.
These tests serve as a filter to refute implausible models. An example would be
to replace the data for the treatment or outcome variable by random data, which is
independent of all other variables. If the model predicts a non-zero causal
effect, it should clearly be refuted.
Other tests include the synthetic addition of random and unobserved common
causes, as well as replacing the original dataset by a subset or a bootstrapped
version of itself.
The idea of such checks is well in line
with the scientific method \cite{popper}: A scientific theory cannot be
proven, but only be falsified. If
all our falsification attempts fail, the theory gains credibility. In the same
sense, we have no means of directly proving our causal model right, so we try
proving it wrong by the refutation tests. A drawback of the method is that
these generic tests might be too weak a filter for distinguishing the correct
model from plausible, but incorrect models. It is our goal to establish
quantitative probing as a general validation framework that provides more
powerful problem-specific refutation tests.

\section{Types of causal models}\label{causal_models}
In order to formulate and evaluate validation strategies for causal models, we
need to specify more clearly what is meant by a causal model. In the context
of validating correlation-based models,
the model can be seen as a black box that answers certain queries. In the
classification setting, the query could be: "Given that we have observed the
following features, what class does the sample belong to?" Similarly, in the
regression setting, we can ask the model: "Given that we have observed the
following features, what is the value of the target variable for this sample?"
Note that the above-presented train/test split strategy for validating these
models does not require any knowledge about what is happening inside of the
black box. We only need to be able to answer the queries for the samples in our
test set, which makes the strategy applicable to a wide range of classification
and regression models.
Following this observation, a strategy for evaluating causal models should also
be applicable to many different variants of causal analysis strategies,
regardless of their implementation details. Therefore, we will use a very
general definition of a causal model: We call everything a causal model that
answers interventional queries, i.e. that estimates probabilities of the form
$p(y|do(x=v))$ to calculate the \emph{average treatment effect (ATE)}
\begin{equation}
  \tau = p(y=1|do(x=1)) - p(y=1|do(x=0))
\end{equation}
from observational data.
For the remainder of the article, we will restrict our studies to these ATEs
in the binary data setting.
Extensions to other types of causal effects, such as conditional average
treatment effects \cite{abrevaya_cate},
natural direct effects and natural indirect effects \cite{causality_pearl} are
possible, but not necessary to illustrate the central ideas of our validation
strategy.
In the same spirit, we refrain from leaving the binary data setting, although
all the arguments readily transfer to more general discrete, categorical and
continuous datasets.

The considered causal model could be
\begin{itemize}
\item[a)] A fully parameterized causal Bayesian network
\cite{causality_pearl, heckerman_causal_bayes_nets},
i.e. a causal graph together
with a conditional probability distribution (CPD)
over each of the $n$ nodes $x_i$,
given its respective parents $\Pi_i$ in the causal graph:
For a given target effect, we set
the treatment variable to a fixed value $v$ and multiply the CPDs
\begin{equation}
  p(x_i | \Pi_i=\pi_i)
\end{equation}
of the non-treatment nodes with the delta distribution for the treatment node $x_t$,
in order to obtain the resulting interventional distribution
\begin{equation}
  p(x_1=v_1, ..., x_n=v_n | do(x_t=v)) =
  \begin{cases}
    \prod_{i\neq t} p(x_i=v_i | \Pi_i=\pi_i) &\text{if } v_t=v \\
    0 &\text{else.}
  \end{cases}
\end{equation}
The interventional mean over the outcome
variable under the given treatment can then be obtained by a marginalization of
the interventional distribution.
\item[b)] The combination of a causal graph, Pearl's do-calculus \cite{causality_pearl}, observational data
and a fixed estimation strategy: The do-calculus allows us to recover an
unbiased statistical estimand for any interventional mean from the causal graph.
This estimand is built purely from do-free expressions, such as ordinary
conditional probabilities, which can subsequently be estimated from the
observational data using the fixed estimation strategy.
\item[c)] The combination of qualitative domain knowledge, a causal discovery
algorithm \cite{glymour_discovery_survey, causal_discovery_survey_vowels},
Pearl's do-calculus, observational data and a fixed estimation
strategy:
The procedure is the same as for method b), except we first need to use the
causal discovery algorithm to recover the causal graph from our qualitative
domain knowledge and the observational data.
\item[d)] Any model that does not rely on graphs for the encoding of qualitative domain knowledge,
e.g. a model that is based on the potential outcomes framework
\cite{rubin_potential_outcomes}. Due to our limited experience with these models, we will focus on
graph-based causal models for the remainder of the article. However, as long as the model can answer
multiple causal queries, it is suitable for validation by quantitative probing.
\end{itemize}

The enumeration shows that causal inference can be a complex pipeline of
multiple analysis steps. Model b) is clearly a special case of model c), where
we assume an omniscient causal discovery algorithm. In the same vein, model a)
is a special case of model b), where we fit the CPDs from our observational
data, assuming that we know the functional form of each CPD.
Applying the do-calculus is no more necessary if the CPDs are available.
In order to consider a general scenario without any restrictive assumptions
about the underlying data generating process,
we introduce the following graph-based causal model type
that covers many simpler types of causal analysis as special cases.

By \emph{causal end-to-end analysis}, we mean the following procedure for a given
dataset and a given target effect.
\begin{itemize}
  \item We preprocess the data by deleting, adding, rescaling or combining variables.
  \item We pass qualitative domain knowledge by specifying which edges must or
  must not be part of the causal graph.
  \item We run a causal discovery algorithm that respects the qualitative domain
  knowledge.
  \item We postprocess the proposed causal graph by deleting, adding, reversing or
  orienting a subset of edges in the causal discovery result.
  \item We identify an unbiased statistical estimand for the target effect and
  validation effects by
  applying the do-calculus to the causal graph.
  \item We estimate the estimands by a method of our choice.
\end{itemize}

\section{Quantitative probing}\label{qprobing}
Validation strategies often assume that we want to predict a single
\emph{target (causal) effect},
once the causal graph has been specified. We will refer to both
the treatment variable and outcome variable of the target causal effect as
\emph{target variables} for ease of notation.
All the other variables in the dataset, which we will call \emph{non-target
variables}, usually are either ignored or treated as confounders, based on the structure
of the causal graph. In either case, they are taken care of by methods like do-calculus
and we do not have to inspect their quantitative causal relationships with any
of the target or non-target variables. However, not taking into
account these \emph{non-target effects} means wasting our domain knowledge:
Just as we can pass parts of the causal graph to a causal discovery algorithm as
a form of qualitative domain knowledge, we can use our expectations about selected
non-target effects as supplementary quantitative domain knowledge.
Analogously to communicating that we  expect an edge between two variables
in the causal graph, we can specifiy that we expect a certain non-target effect
to be close to a given value.
There is a key difference between passing qualitative and quantitative domain
knowledge.
The qualitative domain knowledge can be explicitly accepted as an input by a
causal discovery algorithm, meaning that the procedure
actively uses the knowledge in recovering the correct causal graph from the
observational data \cite{meek_knowledge}.
For quantitative knowledge, on the other hand, it is not even clear where to
pass these desiderata about the causal model and its predictions. Should
we pass them to the estimation procedure and use them for hyperparameter tuning?
Or could it be that the hyperparameters are correct, even when the model fails
to reproduce the expected effect? Think for instance of a failure that
can be attributed to a
misspecification of the causal graph in the preceding discovery step. It is even
conceivable that the expectations have not been met because of a mistake in
data preprocessing, and changing the causal discovery or estimation steps will
lead to an inferior estimate of the target effect. Such considerations make it clear
that the quantitative knowledge can hardly be tied to a specific step in the
end-to-end causal analysis. However, we can always use the knowledge in the
following way: If we are sure that a given non-target effect must come close to a
specific value, but our analysis fails to reproduce this outcome, something must
have gone wrong. We do not know exactly what it is, but we cannot exclude that
the same error has also affected the estimation of our target effect.
Such a failure to reproduce our expectations should therefore
diminish our confidence in our estimate of the target effect. Conversely, if
we specify many different expected non-target effects and all of them are
reproduced by our analysis, our confidence in the estimation of the target
effect by the very same analysis should increase.
As previously mentioned in the discussion of causal validation via refutation
tests \cite{dowhy_paper}, such a strategy is in line with the general logic of
establishing scientific theories \cite{popper}.
Failing to falsify a candidate model by
probing its estimates of non-target effects against previously stated expectations
about their values increases our trust in the model.
Consequently, we will refer to the specified non-target effects as
\emph{quantitative probes} or simply \emph{probes} and to the presented
validation strategy as \emph{quantitative probing}.
In the remainder of this
article, we will present simulation-based evidence for this line of reasoning.

\subsection{Sprinkler example}
\begin{figure} [htb!]
	\begin{center}
		\fbox{\includegraphics[height=8cm]{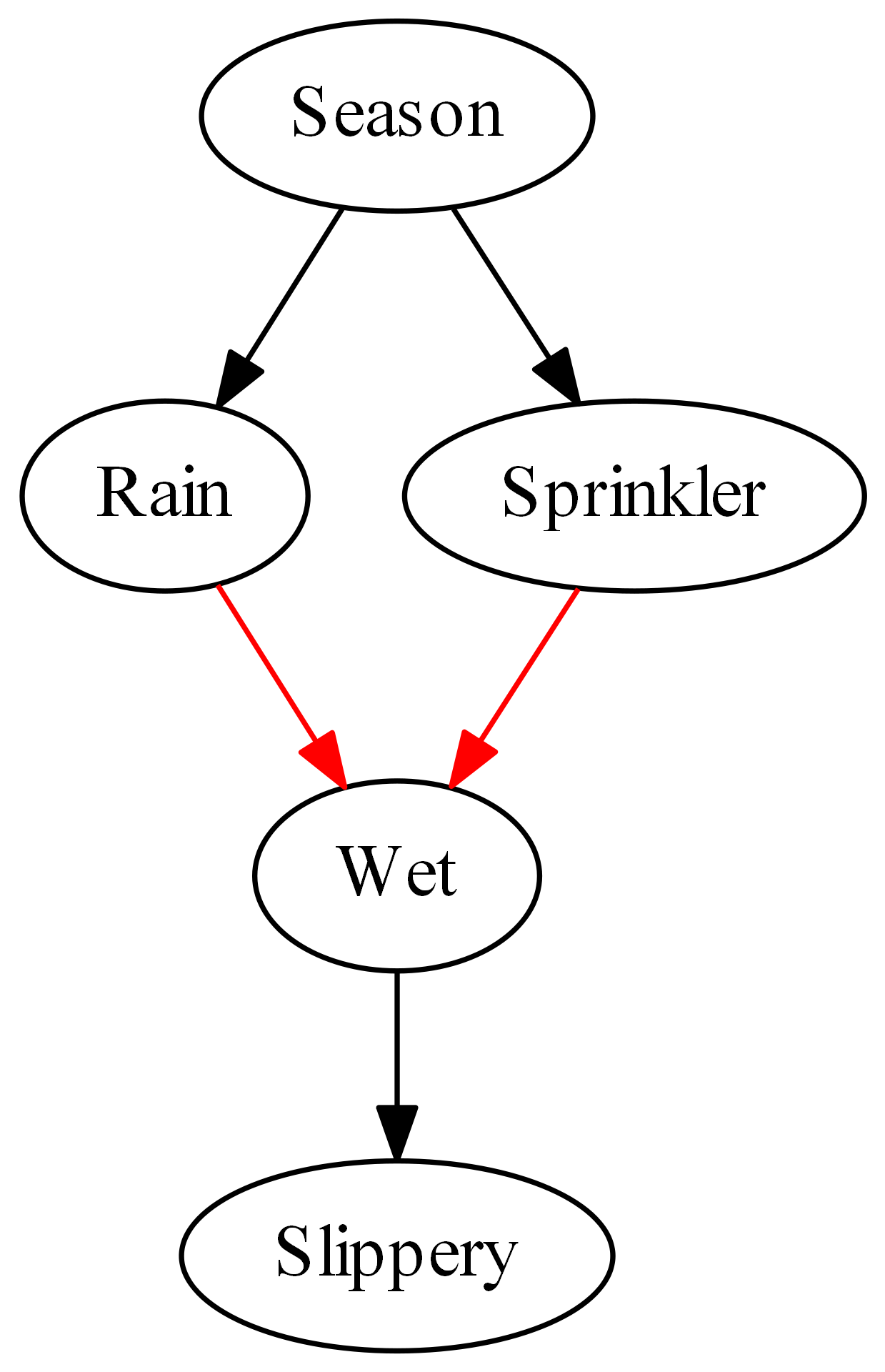}}
		\fbox{\includegraphics[width=8cm]{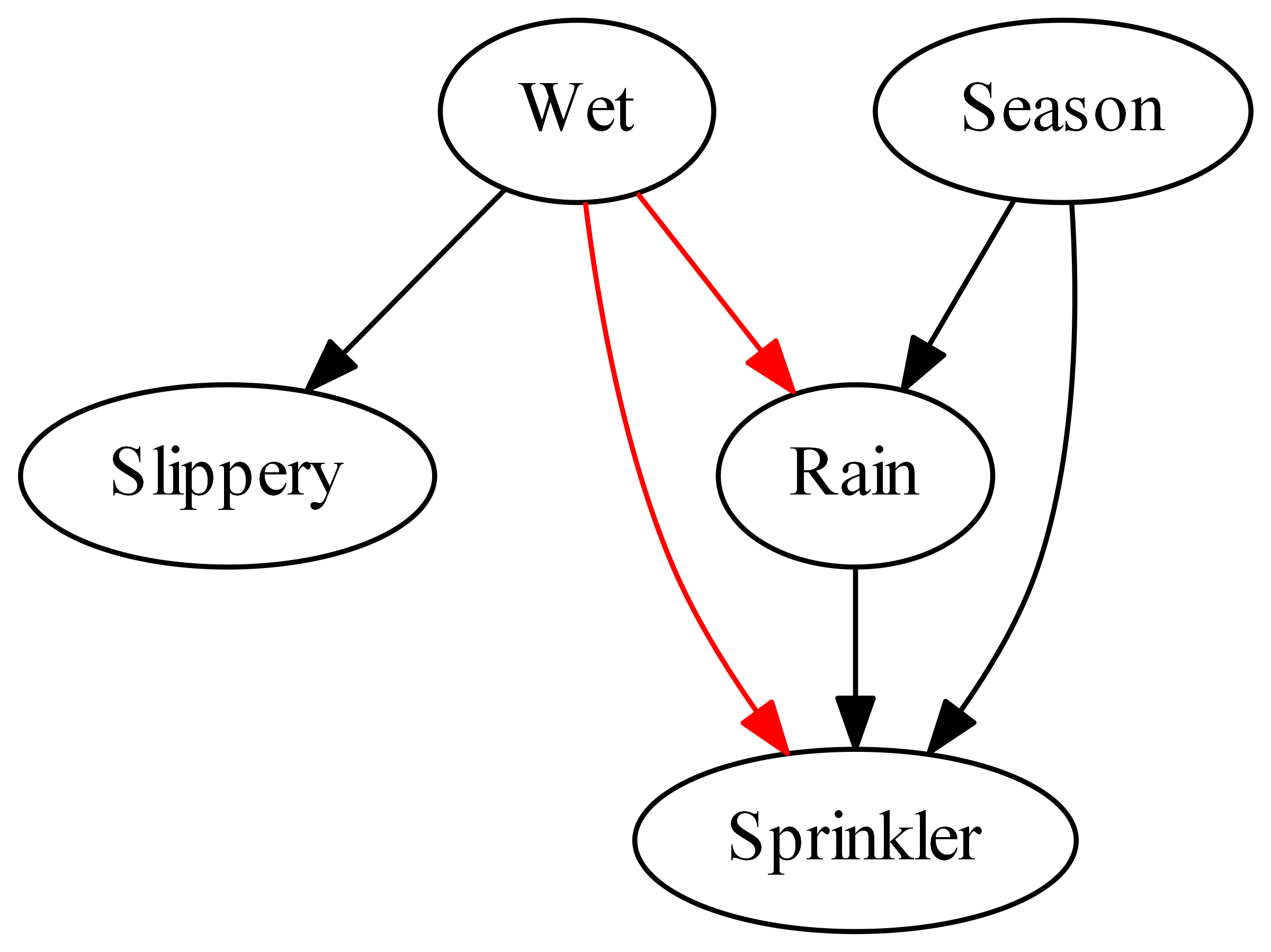}}
		\caption[graph]{
    The recovered graphs correctly (left) or incorrectly (right)
    represent the data
		generating process. Edges prescribed by the respective sets of
    domain knowledge are marked in red,
		the rest of them has been added by the causal discovery algorithm.
    }
		\label{sprinkler}
	\end{center}
\end{figure}

Before we step into the technical discussion, let us illustrate the presented
motivation for quantitative probing using the well-known sprinkler example
\cite{causality_pearl}.
Suppose that we are interested in estimating the ATE of
activating a garden sprinkler on the slipperiness of our lawn.
Estimating this target effect from observational data is the reason why we are
performing the causal end-to-end analysis.
For this hands-on example, we generate data using pgmpy, an open
source Python package for probabilistic graphical models \cite{pgmpy}.
The subsequent analysis is performed using cause2e, an open source
Python package for causal end-to-end analysis \cite{cause2e_github}.
The data consists of $m=10000$ samples, each holding values for $n=5$ variables:
\begin{itemize}
  \item What was the season on the day of the observation?
  \item Was the sprinkler turned on on the day of the observation?
  \item Was it raining on the day of the observation?
  \item Was the lawn wet on the day of the observation?
  \item Was the lawn slippery on the day of the observation?
\end{itemize}
All of the variables are binary, except for the season variable. For simplicity,
we also binarize the season variable in a preprocessing step: We encode 'Winter'
as zero, 'Spring' as one and discard the observations that were made in summer
or autumn.
After preprocessing, the involved variables can no longer change and we want to
leverage our quantitative domain knowledge about them by creating two
quantitative probes:
\begin{itemize}
  \item We expect that turning on the sprinkler will make the lawn wetter, so
  we expect the ATE of 'Sprinkler' on 'Wet' to be greater than zero.
  \item We expect that making the lawn wetter will also make it more slippery,
  so we expect the ATE of 'Wet' on 'Slippery' to be greater than zero.
\end{itemize}
Note that these two probes do not directly imply specific edges in the causal
graph, as the influence could also be mediated via one of the other variables.
In addition to our quantitative domain knowledge, we can now specify qualitative
domain knowledge about the underlying causal graph.
For demonstrative purposes, we choose a configuration that enables
the causal discovery algorithm to recover the causal graph that was used
for generating the data:
\begin{itemize}
  \item We forbid all edges that originate from 'Slippery'.
  \item We forbid all edges that go into 'Season'.
  \item We forbid the edges 'Sprinkler' $\rightarrow$ 'Rain' and 'Season' $\rightarrow$ 'Wet'.
  \item We require the edges 'Sprinkler' $\rightarrow$ 'Wet' and 'Rain' $\rightarrow$ 'Wet'.
\end{itemize}
Note that this configuration still leaves 9 edges whose presence in the graph is
neither required nor forbidden, but needs to be decided by the causal discovery
algorithm.
If we run fast greedy equivalence search \cite{fges}, an optimized version of
the standard
greedy equivalence search \cite{ges}, as a causal discovery algorithm,
we see that we recover the true causal graph
from the data generating process (cf. Figure \ref{sprinkler}).

In the next step, we estimate the target effect and our quantitative probes from the
causal graph and the data via linear regression.
We recover a target effect of $0.52$, but in
principle, we have no idea whether this is a reasonable estimate or not.
However, the analysis shows that the ATE of 'Sprinkler' on 'Wet' is $0.62$ and the ATE of
'Wet' on 'Slippery' is $0.81$. Therefore, our expectations about the known
causal effects have been met by the predictions of the model, which increases
our trust in the model and therefore also in the estimate of the target effect.

As a contrast, let us see how our probes help us detect flawed
causal models: For this purpose, we intentionally communicate wrong assumptions
about the causal graph to the causal discovery algorithm. For each edge that we
have required in the above example, we now require the corresponding reversed
edge. The same applies to the forbidden edges, which are also reversed.
These flawed inputs lead to a
flawed causal discovery result, as can be seen in Figure
\ref{sprinkler} (right).

If we now use the incorrect graph and the data to estimate the three ATE's via
linear regression, we observe a target effect of $0$ because there is no directed
path from 'Sprinkler' to 'Slippery'. The ATE of 'Wet' on 'Slippery' is again
$0.81$ in accordance with our expectations. However, the ATE of 'Sprinkler' on
'Wet' is $0$, even though we expect it to be a positive number. We can use this
failed validation as evidence that something in our end-to-end analysis must
have gone wrong, and therefore the estimate of the target effect should not be
trusted either. In practice, this could lead us to reexamine the qualitative
domain knowledge that we have passed, which would then result in a causal
discovery result that is closer to the true causal graph. Another possibility
would be the examination of the estimation strategy that was used to compute the
numerical estimate from the statistical estimand and the data. In our case, this
would not lead to a correct estimate of the target effect
because the error in the model lies in the
graph, not in the estimation procedure. A third possibility would be a
reexamination of our quantitative expectations, e.g. if we notice that our
expectations were actually about the natural direct effect
instead of the ATE. In any case, formulating
our explicit quantitative knowledge about certain causal effects in the model,
followed by an automated validation at the end of the end-to-end analysis, serves
as a valuable evaluation step that can help us detect modelling errors.

\subsection{Assumptions}\label{assumptions}
In the previous example, we have made several implicit assumptions that are
necessary to leverage the power of quantitative probing for causal model validation.
\begin{itemize}
  \item We need to have quantitative knowledge about some causal effects between
  the observed variables, otherwise we cannot validate the corresponding
  expectations after the analysis.
  However, note that these expectations can be stated with any desired precision:
  We can demand an effect to be simply non-zero, to be positive,
  to be above a certain threshold or
  even to be situated within a narrow neighborhood of an exact target value.
  \item Estimating the probes should not be excessively complicated to perform
  in addition to the estimation of the target effect. In our example, we could
  reuse the causal graph that we had already constructed for target effect
  estimation, in order to identify unbiased statistical estimands for the
  probes. The estimation itself only required fitting linear regression models
  and reading off the respective coefficients for each probe.
  In a setting where the estimation of the probes is more costly,
  the benefit of using quantitative probing could be
  overshadowed by the required additional effort.
  \item The parts of the data generating process that are responsible for the
  target effect must be in some way related to the parts that are responsible
  for the probes. An example would be that all variables in the target effect
  and in the probes belong to
  the same connected component of the causal graph. Otherwise, our model might
  be perfectly accurate for the component that holds all the probes,
  but flawed in the separate component that produces the target effect.
\end{itemize}

\section{Simulation study}\label{results}
In this chapter, we provide experimental backing for the concept of quantitative
probing as a method of validating causal models.

\subsection{Simulation setup}
In order to have access to both a ground truth and easy parameterization of the
experiments, we chose a setup consisting of the repeated
execution and evaluation of the following parameterized simulation run:

\begin{itemize}
  \item Choose $n$ (number of nodes), $p_{edge}$, (edge probability),
  $m$ (number of samples),
  $p_{hint}$ (hint probability),
  $p_{probe}$ (probe probability) and
  $\epsilon_{probe}$ (probe tolerance).
  \item Draw a random directed acyclic graph (DAG)
  with $n$ nodes $x_1,...,x_n$. Random means that for each of the $n^2$ possible
  directed edges, we include the edge with a probability $p_{edge}$. After all the
  edges have been selected, check whether the result is a DAG.
  If not, repeat the procedure.
  \item Draw a random binary CPD for each node $x_i$. The
  entries $p(x_i=1|\Pi_i=\pi_i)$, which fully determine the CPD, are sampled
  from a uniform distribution on $[0,1]$.
  \item Draw $m$ samples from the resulting joint distribution over
  $(x_1, ..., x_n)$.
  \item Select a proportion $p_{hint}$ of all the edges (rounded down) in the
  causal graph and
 add their presence to the qualitative domain knowledge.
  \item Randomly choose a nontrivial target effect and $p_{probe} \cdot n^2$
  (rounded down)
  other
  treatment-outcome pairs that will serve as quantitative probes. By nontrivial,
  we mean that there exists a directed path from the treatment to the outcome in
  the causal graph, because otherwise any causal effect is trivially zero.
  \item Calculate the corresponding ATEs for the target effect and the probes
  from the fully specified causal Bayesian network, in order to obtain a ground truth.
  \item Run a causal end-to-end analysis, using the $m$ observational samples
  and the qualitative domain knowledge, and report the discovered causal
  graph, the estimate of the target effect and the hit rate for the quantitative
  probes. The hit rate is defined as the proportion of probes that have been
  correctly recovered by the analysis.
  In order to account for numerical errors and statistical fluctuations,
  we allow an absolute deviation of $\epsilon_{probe}$ from the true
  value for a probe estimate to be considered successful.
  \item Report the number of edges that differ between the true and the
  discovered graph, as well as the absolute and relative error of the
  target effect estimate.
\end{itemize}

In this article, we report the results for experiments with
$n=7$ nodes,
an edge probability of $p_{edge}=0.1$,
$m=1000$ samples per DAG,
$p_{hint}=0.3$, meaning
that we suppose that we know $30$\% (rounded down) of the correct causal edges,
$p_{probe}=0.5$, meaning that we use half of the possible causal effects as
quantitative probes, and
$\epsilon_{probe} = 0.1$, meaning that we consider probe estimates
successful if they deviate no more than $0.1$ from the true value on an
additive scale.
As in the previous example, the causal discovery was performed using fast greedy equivalence search
\cite{fges} and all ATEs were estimated using linear regression.
Our hypothesis is that an end-to-end analysis that results in a high hit rate
for the quantitative probes is more likely to have found both the true causal
graph and the true target effect.

\subsection{Used software}
The programmatic implementation of the simulation relies on several open-source
Python packages:
At the beginning of the pipeline, the networkx package \cite{networkx}
was used for sampling DAGs based on the edge probability $p_{edge}$.
The pgmpy package \cite{pgmpy} enabled us to build a probabilistic graphical model from the
given DAG by adding random CPDs for each node.
The resulting model was then used to sample data for creating the observational
dataset, as well as for obtaining the true ATEs by simulating data from
interventional distributions.
The causal end-to-end analysis and the validation of the quantitative probes
were executed with the help of cause2e \cite{cause2e_github}.
All plots were generated using Matplotlib \cite{matplotlib}.
In order to make the setup reusable, easily parametrizable and open for
extension by other researchers, an open-source Python package for quantitative
probing was developed around the above software setup (see Chapter \ref{code}).

\subsection{Results}
\begin{figure} [htb!]
	\begin{center}
		\includegraphics[width=6cm]{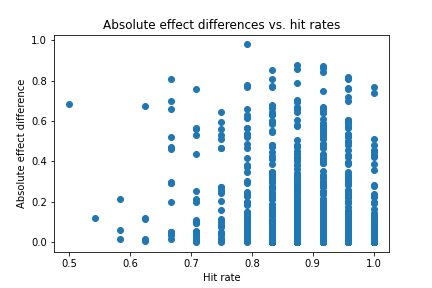}
		\includegraphics[width=6cm]{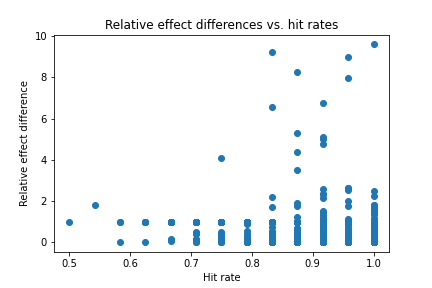}
		\includegraphics[width=6cm]{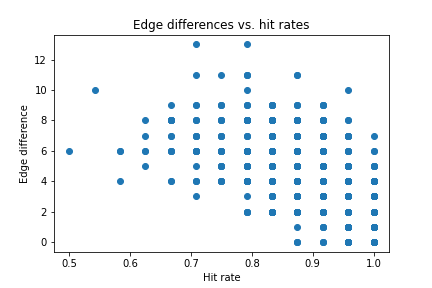}
		\includegraphics[width=6cm]{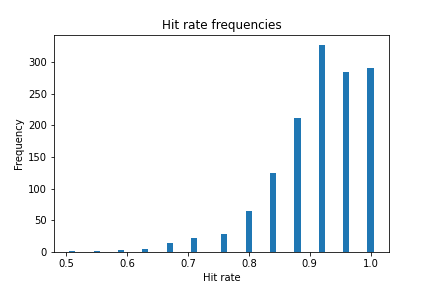}
		\caption[absolute_effect_differences]{
    The top row shows plots of the
    absolute (left) and
    relative (right) differences between the true target effect and the estimated
    result against the hit rate.
    The bottom row shows a plot of the structural hamming distance
		between the true causal graph and the causal discovery result against the
		hit rate (left), as well as a histogram of the observed hit rates (right).
    }
		\label{single}
	\end{center}
\end{figure}

In order to support the above hypothesis, we plot
the aggregated results of 1378 runs.
Figure \ref{single} shows the information for every single run as a separate
data point: The $x$-coordinate indicates the hit rate of the run in each of the
four plots, whereas the $y$-coordinate describes varying quantities of
interest:
\begin{itemize}
\item The plot to the upper left shows the absolute difference between the
estimated and the true value of the target effect.
\item The plot to the upper right shows the relative difference
$\lvert \frac{\hat{\tau} - \tau}{\tau} \rvert$
between the estimated value~$\hat{\tau}$
and the true value~$\tau$ of the target effect.
Note that no division by zero happens, since all target effects were chosen to
be nontrivial.
\item The plot to the lower left shows the structural hamming distance
\cite{tsamardinos_structural_hamming_distance}
 between the
true and the discovered graph.
This includes both, edges that are present
in only one of the graphs, as well as reversed edges.
\item The plot to the lower right is the only aggregation:
It shows the absolute frequencies of the hit rates over all runs.
\end{itemize}

\begin{figure} [htb!]
	\begin{center}
		\includegraphics[width=6cm]{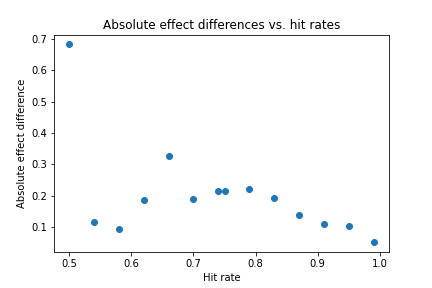}
		\includegraphics[width=6cm]{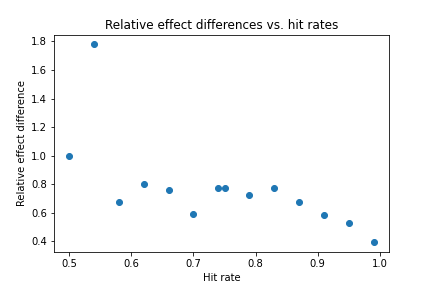}
		\includegraphics[width=6cm]{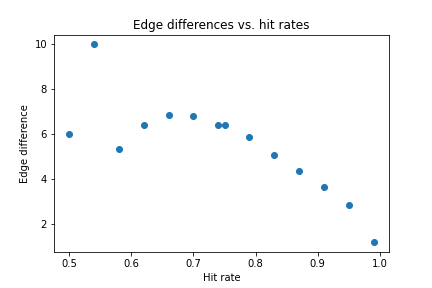}
		\includegraphics[width=6cm]{plots/hit_rate_frequencies.png}
		\caption[absolute_effect_differences]{
    Aggregated results:
    The top row shows plots of the mean
    absolute (left) and
    relative (right) difference between the true target effect and the estimated
    result against the hit rate.
    The bottom row shows a plot of the mean structural hamming distances
		between the true causal graph and the causal discovery result against the
		hit rate (left), as well as a histogram of the observed hit rates (right).
    }
		\label{unconnected}
	\end{center}
\end{figure}

As we can see, the data in the upper plots does not seem to support our
hypothesis at all: We expect to see the points following a trend from the upper
left (few successful probes, large estimation error) to the lower right (many
successful probes, small estimation error), but there seems to be no downward
trend. On the contrary, Figure \ref{single} (upper left) even shows a higher
number of significant estimation errors when the hit rate is high.
At least the second concern can be resolved by a look at the hit rate
frequencies in Figure \ref{single} (lower right): Almost all of the runs show a
hit rate of at least $0.8$. The high ratio of successful probes is likely
connected to
the fact that many of the randomly chosen treatment-outcome-pairs in the probes
are not connected by a directed path in the true graph.
Even if the discovered graph is not
completely correct, it is sufficient not to introduce a directed path by mistake,
in order to estimate the probe successfully. Therefore, the higher number of
significant relative estimation errors in the plot can be explained by the
higher number of data points for the corresponding hit rates.
Similarly, the above-mentioned apparent lack of a downward trend is simply due
to visualization problems caused by the high number of data points.
In order to resolve this issue, we replace the many single data points for each
hit rate by one data point whose $y$-coordinate is the mean value over the
plotted quantity for the given hit rate. The results are shown in Figure
\ref{unconnected}: Since the plots are derived from the same runs as before, the
hit rate histogram in Figure \ref{unconnected} (bottom right) is unchanged. The
other three plots now show the expected downward trend for the regions that
contain a sufficient number of samples, indicating that runs
with a higher hit rate performed better at estimating the target effect and
recovering the causal graph from data and domain knowledge.
It even seems that the relationship between the hit rate and the other variables is
approximately linear, but a theoretical foundation for this observation could
not be established.

\subsection{Outlier analysis}\label{outlier_analysis}
The results indicate that the probability of having recovered both, the true
causal graph and the correct target effect, from observational data and domain
knowledge increases with the amount of correctly estimated quantitative probes, i.e. the hit rate.
However, the presented evidence only supports this in a probabilistic manner and
even a perfect estimation of all probes does not guarantee a successful causal
analysis. This contrast is reflected in the above plots:
On the one hand, the mean errors in Figure
\ref{unconnected} approach $0$, as the hit rate approaches
$1$.
On the other hand, the non-aggregated plots in Figure \ref{single} show numerous
data points with a hit rate of $1$ and considerable errors in both, graph and
target effect recovery.
More precisely,
our data contains $14$ runs that simultaneously have a perfect hit rate of $1$
and an absolute estimation error of at least $0.2$.
In order to understand the thereby evidenced limitations of the quantitative
probing approach, we look at some of these outliers more closely.

\subsubsection{Connectivity}

\begin{figure} [htb!]
	\begin{center}
    \fbox{\includegraphics[height=6cm]{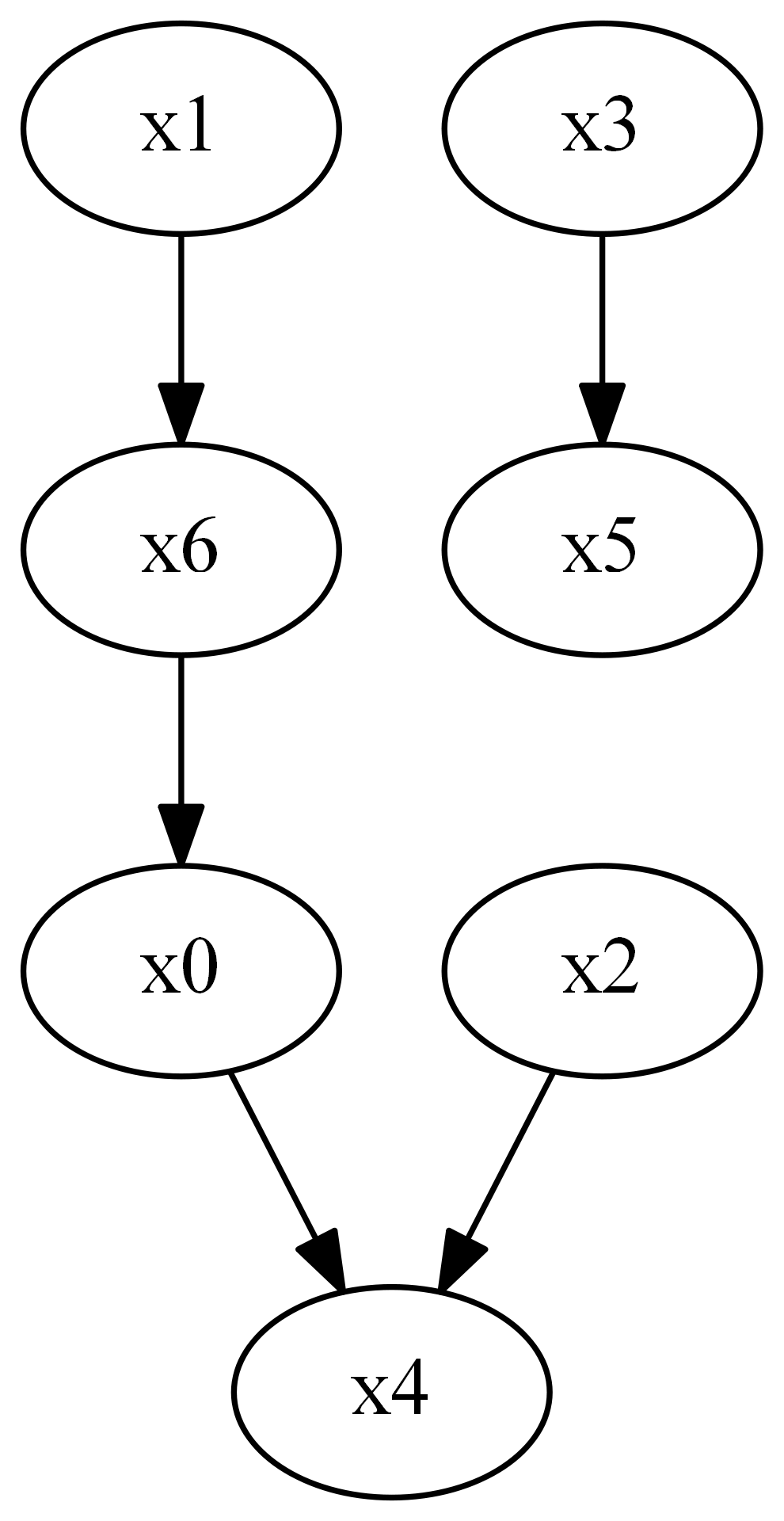}}
		\fbox{\includegraphics[height=6cm]{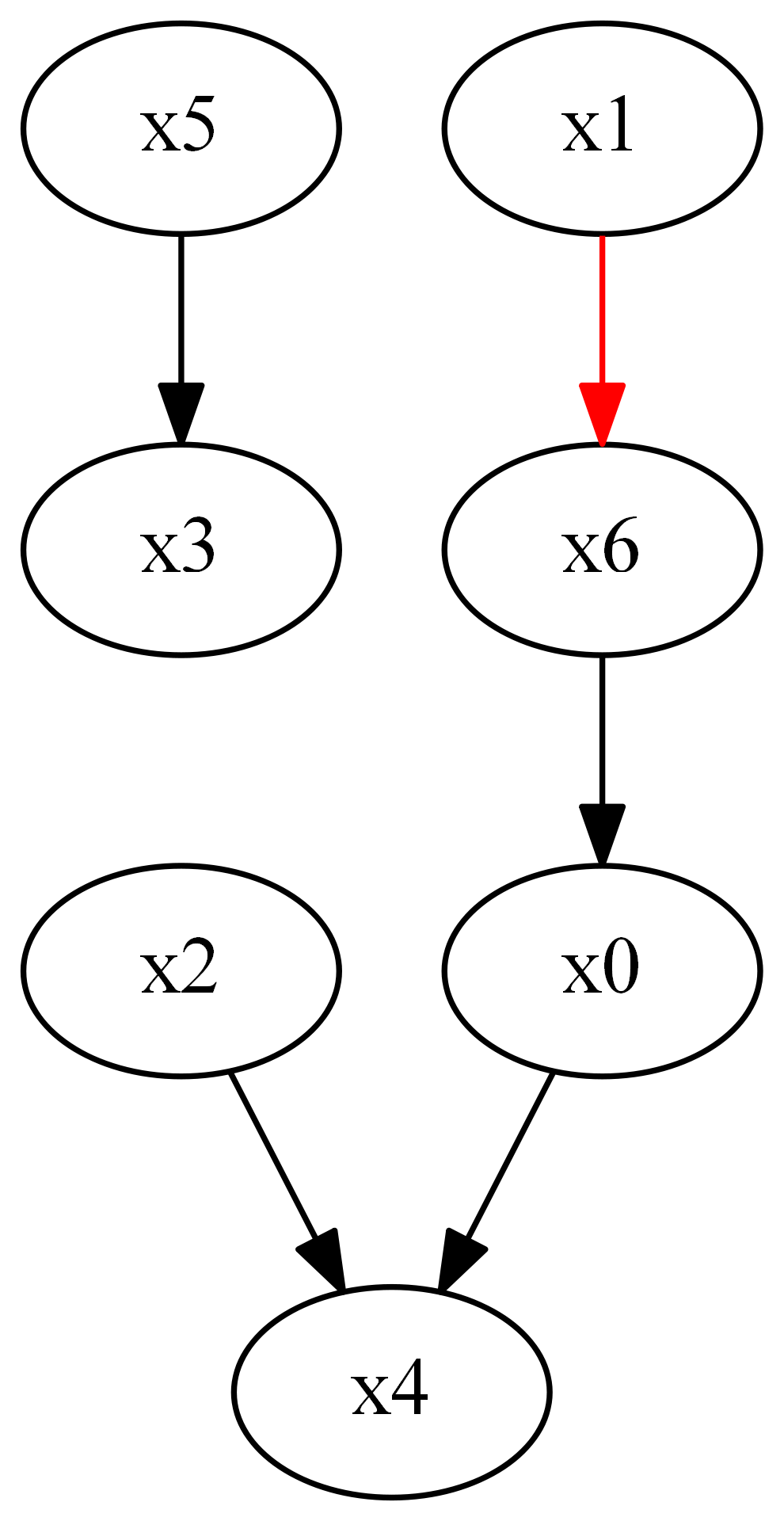}}
		\caption[absolute_effect_differences]{
    Plot of the true (left) and
		discovered (right) causal graphs for an outlier run with treatment
    $x_3$ and outcome $x_5$.
    In the discovered graph, the red edge
    $x_1 \rightarrow x_6$
    has been required by domain knowledge.
    The edge
		$x_3 \rightarrow x_5$ has been oriented incorrectly.}
		\label{outlier}
	\end{center}
\end{figure}

Consider the run depicted in Figure \ref{outlier}:
The target effect was the ATE of $x_3$ on $x_5$, which has
been incorrectly estimated to be $0$ instead of $0.5$, although all of the
probes have been correctly estimated.
An examination of the graph structures immediately explains the phenomenon.
The true graph and the
discovered graph are identical, except for one edge between $x_3$ and $x_5$,
which has been reversed by the causal discovery algorithm. This is not
surprising, as the two structures are Markov equivalent and could only have been
discerned by passing the correct orientation of the edge as domain knowledge.
However, the only edge included in the domain knowledge is the edge between
$x_1$ and $x_6$ (red). As a result, all of the probes have been estimated
correctly, leading to a perfect hit rate of $1$. Of course, the perfect
performance in one connected component of the causal graph has no benefits for a
task that relates only to the other component of the graph, such as the
estimation of our target causal effect.
These findings illustrate that it is not only important to correctly recover the
probes, but also to select helpful probes in the first place.
Given that we have not enforced any connectivity constraints during DAG
generation, it is plausible that the proposed validation technique has
encountered problems in graphs with multiple connected components.

\begin{figure} [htb!]
	\begin{center}
		\includegraphics[width=6cm]{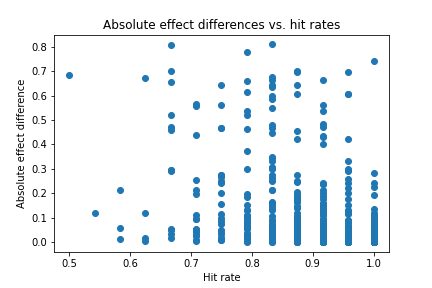}
		\includegraphics[width=6cm]{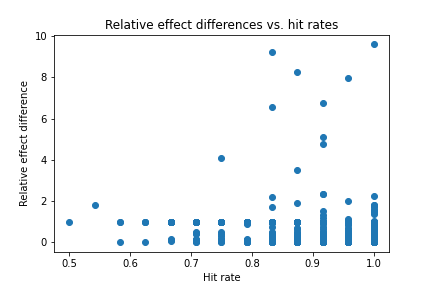}
		\includegraphics[width=6cm]{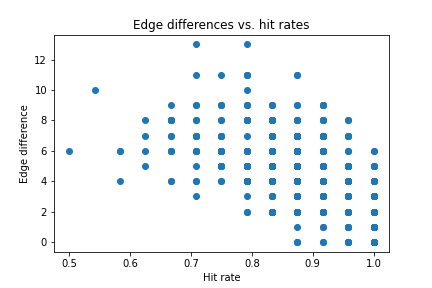}
		\includegraphics[width=6cm]{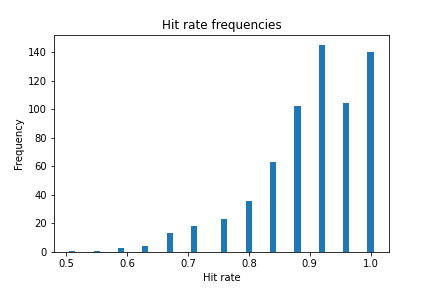}
		\caption[absolute_effect_differences]{
    Results for all runs with connected causal graphs:
    The top row shows plots of the
    absolute (left) and
    relative (right) differences between the true target effect and the estimated
    result against the hit rate.
    The bottom row shows a plot of the number of structural hamming distances
		between the true causal graph and the causal discovery result against the
		hit rate (left), as well as a histogram of the observed hit rates (right).
    }
		\label{connected_single}
	\end{center}
\end{figure}

\begin{figure} [htb!]
	\begin{center}
		\includegraphics[width=6cm]{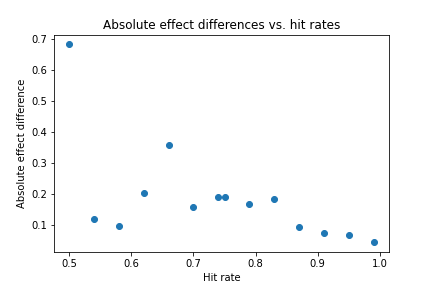}
		\includegraphics[width=6cm]{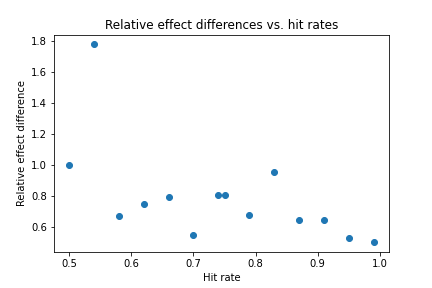}
		\includegraphics[width=6cm]{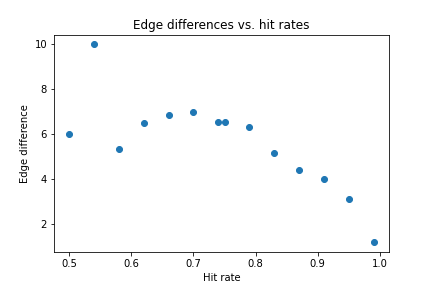}
		\includegraphics[width=6cm]{plots/connected_hit_rate_frequencies.png}
		\caption[absolute_effect_differences]{
    Aggregated results for all runs with connected causal graphs:
    The top row shows plots of the mean
    absolute (left) and
    relative (right) difference between the true target effect and the estimated
    result against the hit rate.
    The bottom row shows a plot of the mean structural hamming distances
		between the true causal graph and the causal discovery result against the
		hit rate (left), as well as a histogram of the observed hit rates (right).
    }
		\label{connected}
	\end{center}
\end{figure}

In order to confirm these explanations, we filter out all experiment runs where
the true causal graph consists of more than one connected component, and recreate Figures
\ref{single} and \ref{unconnected} from the reduced dataset of 653 runs.
The resulting Figures \ref{connected_single} and \ref{connected} indeed show
fewer deviations from the bottom right, indicating that a sizeable proportion of the
runs where our validation approach failed were linked to the problem of
disconnected graphs. If we apply the above outlier filter, we are left with only
$4$ runs with perfect hit rate and an absolute estimation error of over $0.2$.
In practice, this suggests that we should choose probes that we suspect to be in
the same component of the causal graph as the target variables.
It seems plausible that even within one component, the probes that are closer to
the target effect will be more useful for judging the correctness of the model.

\subsubsection{Probe coverage}
\begin{figure} [htb!]
	\begin{center}
    \fbox{\includegraphics[height=6cm]{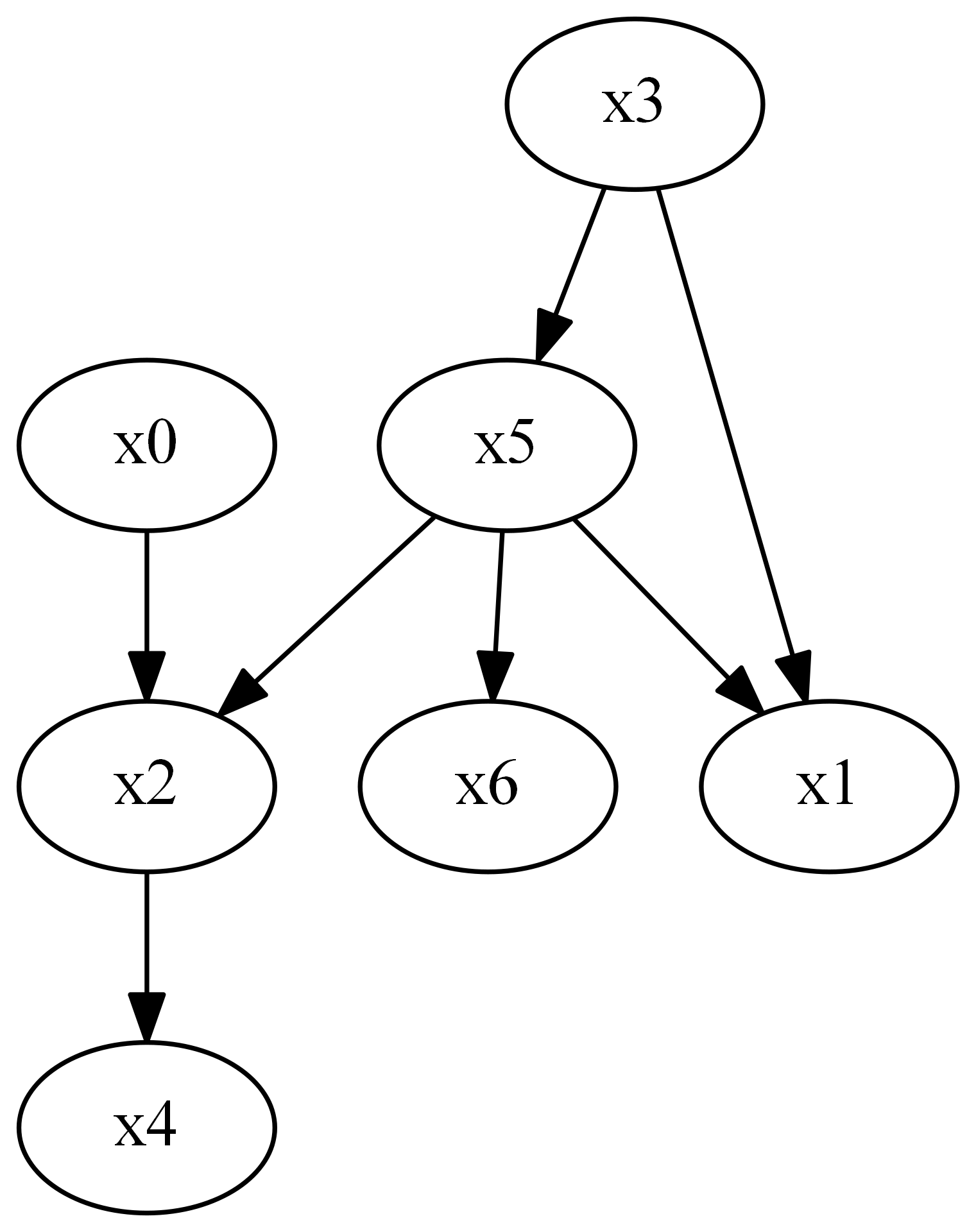}}
		\fbox{\includegraphics[height=6cm]{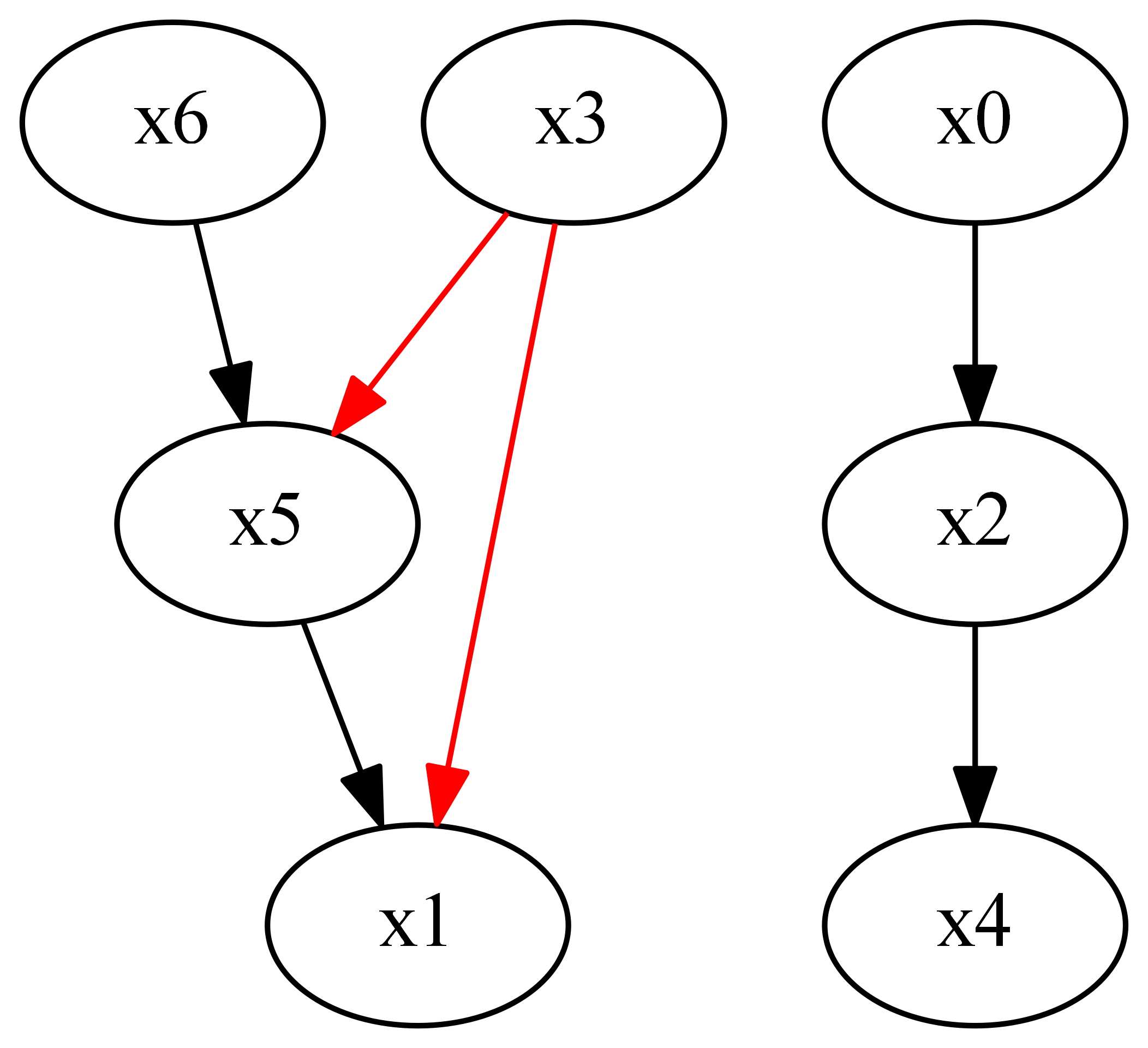}}
		\caption[absolute_effect_differences]{
    Plot of the true (left) and
		discovered (right) causal graphs for an outlier run with treatment
    $x_5$ and outcome $x_6$.
    In the discovered graph, the red edges
    $x_3 \rightarrow x_5$ and $x_3 \rightarrow x_1$
    have been required by domain knowledge.
    The edge
		$x_5 \rightarrow x_6$ has been oriented incorrectly and the edge $x_5
		\rightarrow x_2$ is missing.
    }
		\label{connected_outlier_coverage}
	\end{center}
\end{figure}

In order to understand the remaining outliers, we offer two explanations:
The first one is related to the probe coverage. Given that in most applications,
it is unrealistic to assume that the analyst knows all causal effects except for
the target effect, we have selected only $24$ of the $7 \cdot 7 = 49$ causal
effects as probes. This suggests that some of the erroneous analyses in the
outlier runs could have been captured by increasing the number of probes,
as is illustrated in Figure \ref{connected_outlier_coverage}:
The estimation of the ATE of $x_5$ on $x_6$ has yielded a
result of $0$ instead of the true value $0.28$, although all the probes have
been correctly estimated. A closer look at the 24 probes reveals that the
ATE of $x_3$ on $x_6$ has not been used as a probe. This probe could have
detected the incorrect graph, since its effect is trivially zero in the
discovered graph, but non-zero in the true graph.

\subsubsection{Probe tolerance}

The second explanation focusses on the definition of the hit rate, which is
clearly a deciding factor for marking a run as an outlier. The hit rate is high
if many probes have been correctly estimated, and "correctly" means that the
estimate has to lie within some reasonable bounds around the true effect.
While this seems straightforward, the problem lies in the definition of the
bounds:
Should we specify an absolute error margin that applies to all of the probes? Or
should we specify a relative margin that depends on the size of each of the
probe effects?
Using an absolute margin of $\epsilon_{probe}=0.1$ to both sides, as we did in our
experiments, can be a good fit for a true effect size of $1$, but it might be
dangerous for a true effect size of $0.001$ (underreject) or $1000$
(overreject).

\begin{figure} [htb!]
	\begin{center}
    \fbox{\includegraphics[height=6cm]{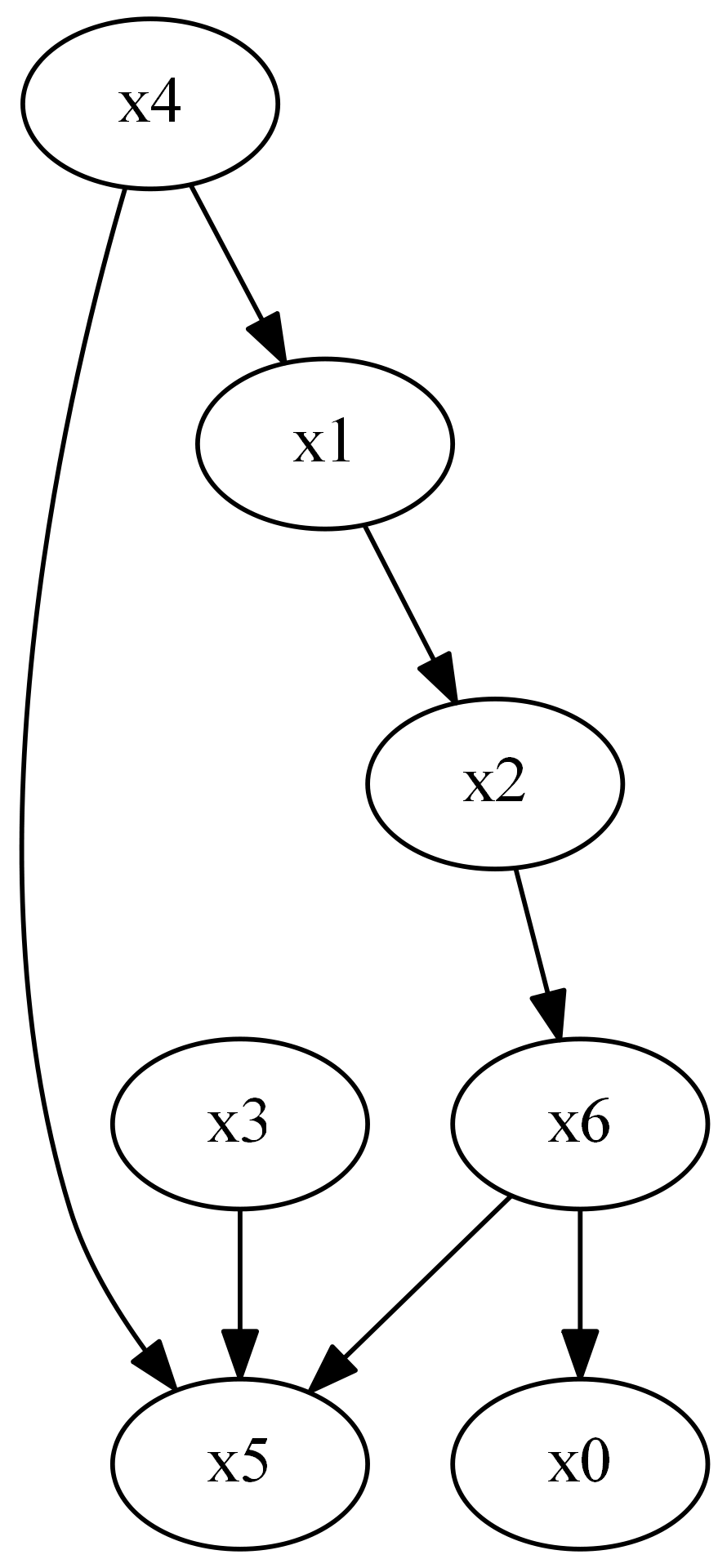}}
		\fbox{\includegraphics[height=6cm]{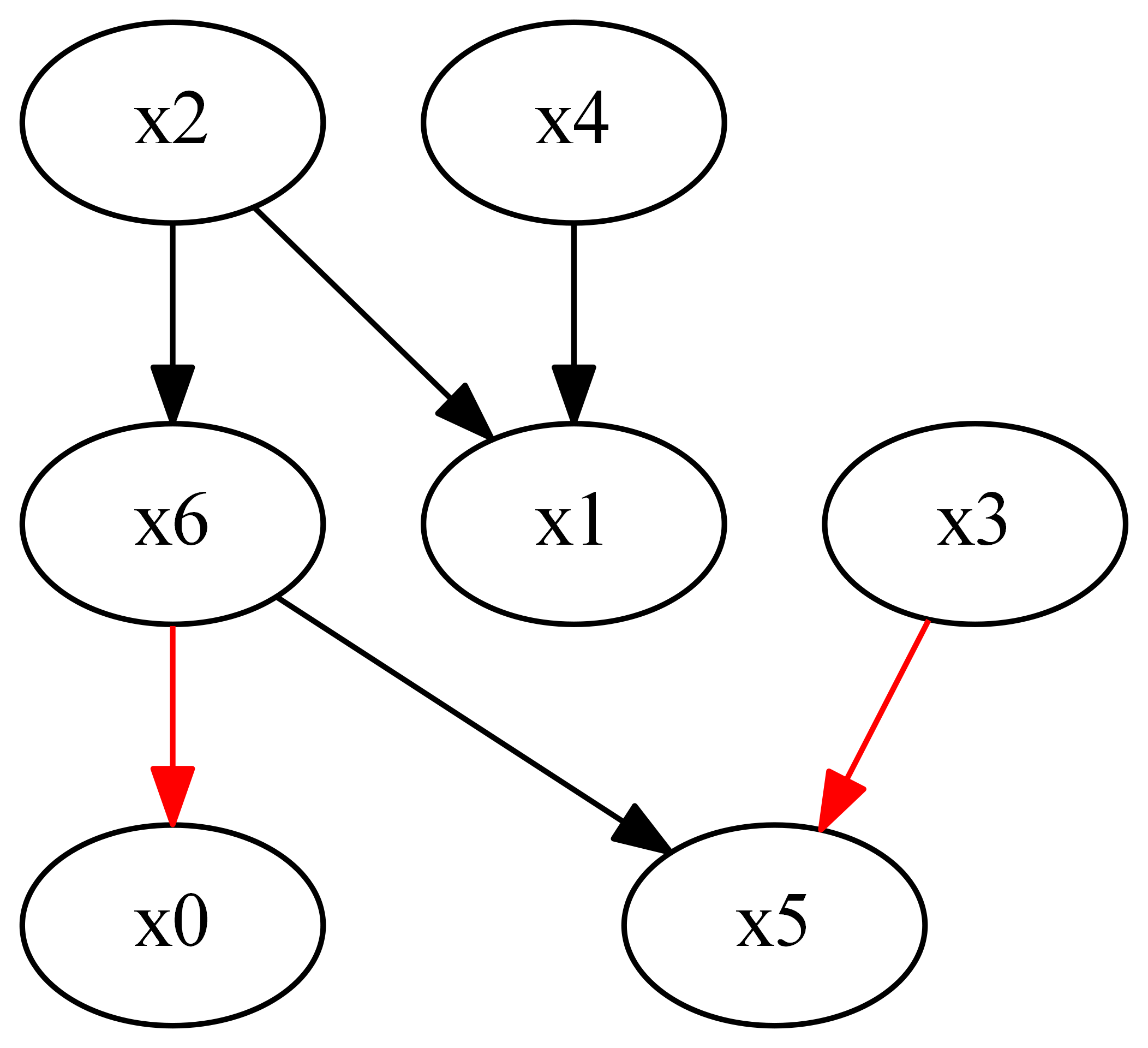}}
		\caption[absolute_effect_differences]{
    Plot of the true (left) and
		discovered (right) causal graphs for an outlier run with treatment
    $x_1$ and outcome $x_2$.
    In the discovered graph, the red edges
    $x_6 \rightarrow x_0$ and $x_3 \rightarrow x_5$
    have been required by domain knowledge.
    The edge
		$x_1 \rightarrow x_2$ has been oriented incorrectly and the edge $x_4
		\rightarrow x_5$ is missing.
    }
		\label{connected_outlier_tolerance}
	\end{center}
\end{figure}

To illustrate this line of thought, we look at the run in Figure
\ref{connected_outlier_tolerance}:
The ATE of $x_1$ on $x_2$ was erroneously estimated to be $0$ instead of
$-0.24$, although all of the probes were estimated correctly.
In this case, it is remarkable that the ATE of $x_4$ and $x_2$ was one of the
probes. This probe is trivially $0$ in the discovered graph, as there is no
directed path from $x_4$ to $x_2$. In the true graph, however, it can only
vanish in the degenerate case. Indeed, the true effect is $0.07$. Given our
generic bounds determined by $\epsilon_{probe}=0.1$, the incorrect estimate of
$0$ falls within the acceptance interval $[-0.03, 0.17]$ and the error goes
unnoticed. This could have been avoided by specifying proper bounds for each of
the probes, a task whose feasibility in practice depends on the available domain
knowledge.
It is worth noting that all the inspected outliers either show a true or an
estimated target effect that is trivially $0$. The absence of more subtle estimation
problems where the directed path exists in both, the true and the recovered
graph, but the estimation is jeopardized by a difference in backdoor
paths, is probably due to the low number of variables in the generated DAGs.
However, quantitative probing is applicable without any modifications to more
complicated graph structures.

\section{Conclusion and outlook}\label{conclusion}
In summary, we have introduced the method of quantitative probing for validating
causal models.
We identified the additional difficulties in validating causal models when
compared to traditional correlation-based machine learning models by reviewing the role of the
i.i.d. assumption.
In analogy to the model-agnostic train/test split, we
proposed a validation strategy that allows us to treat the underlying causal model as a black
box that answers causal queries.
The strategy was put into context as an extension of the already existing technique of using
refutation checks, which is in line with the established logic of scientific discovery.
After illustrating the motivation behind the concept of
quantitative probing using Pearl's sprinkler example, we presented and discussed the results of
a thorough simulation study.
While being mostly supportive of our hypothesis that quantitative probes can
be used as an indicator for model fitness, the study also revealed shortcomings of the method, which
were further analyzed using exemplary failing runs.

To conclude the article, we revisit the assumptions from Chapter \ref{assumptions} and
the ideas from Chapter \ref{outlier_analysis},
in order to identify topics for future research.
\begin{itemize}
  \item Is there a way to explain the roughly linear association between the hit
  rate and the edge/effect differences in Figure \ref{unconnected}?
  Answering this question would provide a solid theoretical backing for the
  quantitative probing method, in addition to the simulation-based evidence
  presented in this article.
  \item Can we quantify the "usefulness" of each employed quantitative probe for
  detecting wrong models?
  An example would be to plot the results for a single fixed hit rate, but each
  data point could represent runs that used only specific probes, e.g. only those
  whose variables lie within a maximum distance from the target variables.
  What other criteria could make a specific probe useful?
  Answering this question would aid practitioners in eliciting specific
  quantitative knowledge from domain experts.
  \item Given the probe coverage issue in Figure
  \ref{connected_outlier_coverage},
  is there a way to model how the effectiveness of the method depends on
  the number of used quantitative probes?
  Answering this question would aid practitioners in gauging the amount of
  required quantitative domain knowledge for model validation.
  \item Given the probe tolerance issues in Figure
  \ref{connected_outlier_tolerance},
  how much more useful do the probes become for detecting wrong models if we
  narrow the allowed bounds in the validation effects?
  How can we determine ideal bounds to avoid overreject and underreject?
  Answering this question would provide important guidance to both researchers
  and practitioners, as the bounds must be actively chosen by the investigator
  in each application of quantitative probing.
\end{itemize}

Similarly to how qualitative knowledge by domain experts
can assist causal discovery procedures in recovering the
correct causal graph from observational data,
we believe that quantitative probing can serve as a natural method of incorporating quantitative
domain knowledge into causal analyses.
By answering the above questions, the proposed validation strategy can evolve into a tool that
builds trust in
causal models, thereby facilitating the adoption of causal inference techniques in various
application domains.

\section{Acknowledgments}
We are grateful for the support of our colleagues at ams OSRAM and the University of Regensburg.
Sebastian Imhof helped sharpen the idea of using known causal effects for model validation during
multiple fruitful conversations.
Other validation approaches for causal models were compared in several meetings of our causal
inference working group \cite{working_group}.
The idea of different degrees of usefulness for different types of quantitative probes
was brought up in a
discussion with members of the Department of Statistical Bioinformatics at the University of
Regensburg.
Heribert Wankerl deserves special credit for proofreading the paper.

\section{Code and data availability}\label{code}
All of the above results can be reproduced by running two notebooks in the companion GitHub
repository of this article,
which also contains the open-source qprobing Python package
\cite{qprobing_github}.
The qprobing package relies on the open-source cause2e Python package for performing the causal
end-to-end analysis, which is hosted in a separate GitHub repository
\cite{cause2e_github}.

\bibliographystyle{unsrt}
\bibliography{qprobing}

\end{document}